\documentclass{article}
\usepackage{iclr2027_arxiv,times}
\usepackage{etoolbox}
\makeatletter
\patchcmd{\@maketitle}{\rule{\z@}{24pt}}{\rule{\z@}{18pt}}{}{}
\patchcmd{\@maketitle}{\rule{\z@}{24pt}}{\rule{\z@}{18pt}}{}{}
\patchcmd{\@maketitle}{\rule{\z@}{24pt}}{\rule{\z@}{18pt}}{}{}
\makeatother

\ifdefined\pdfinfoomitdate\pdfinfoomitdate=1\fi
\ifdefined\pdftrailerid\pdftrailerid{}\fi
\ifdefined\pdfsuppressptexinfo\pdfsuppressptexinfo=15\fi

\usepackage{amsmath,amsfonts,bm}

\def\eqref#1{equation~\ref{#1}}

\def\1{\bm{1}}

\DeclareMathAlphabet{\mathsfit}{\encodingdefault}{\sfdefault}{m}{sl}
\SetMathAlphabet{\mathsfit}{bold}{\encodingdefault}{\sfdefault}{bx}{n}

\usepackage{xurl}
\usepackage{microtype}
\usepackage{graphicx}
\usepackage{amsmath}
\usepackage{booktabs}
\usepackage{colortbl}
\usepackage{multirow}
\usepackage{arydshln}
\usepackage{algorithm}
\usepackage{algorithmic}
\usepackage{hyperref}
\hypersetup{pdfauthor={},pdfcreator={},pdfproducer={}}

\AddToHook{env/table/begin}{\setlength{\belowcaptionskip}{6pt}}
\AddToHook{env/table*/begin}{\setlength{\belowcaptionskip}{6pt}}
\graphicspath{{Figures/}}

\newcommand{\method}{\textsc{AMD}}
\newcommand{\biregime}{\textsc{Fixed-Bi}}
\newcommand{\uniregime}{\textsc{Fixed-Uni}}
\newcommand{\autoregime}{Adaptive}
\newcommand{\mainref}[1]{\ref{#1}}

\title{Adaptive Mutual Distillation for Balanced\\
Multi-Task Post-Training of Large Language Models}

\iclrfinalcopy
\author{
Baohang Li$^{1}$ \And Xiaocheng Feng$^{1,2}$ \And Yichong Huang$^{1}$ \And Chengpeng Fu$^{1}$ \AND
Wenshuai Huo$^{1}$ \And Zekun Zhou$^{1}$ \And Zekun Yuan$^{1}$ \And Tingjia Zhang$^{1}$ \And Bing Qin$^{1}$
\AND
\textnormal{\small $^{1}$Harbin Institute of Technology \qquad $^{2}$Peng Cheng Laboratory}\\[-2pt]
\textnormal{\small\texttt{baohangli@ir.hit.edu.cn}}
}

\begin{document}

\maketitle

\begin{abstract}
Multi-task post-training of large language models (LLMs) aims to improve performance across tasks with unequal amounts of training data. Existing methods focus primarily on balancing task contributions during single-model training. Different task-balancing strategies can produce models with complementary strengths, creating opportunities for mutual distillation. However, the usefulness of cross-model supervision can vary across tasks, transfer directions, and stages of training. We propose \textbf{Adaptive Mutual Distillation (AMD)}, a collaborative post-training framework that jointly trains two models with different task-balancing strategies. AMD evaluates candidate adjustments to distillation weights through short training probes shared across tasks, then uses task-wise validation scores to select an adjustment for each task and transfer direction. Across six benchmarks and three LLM backbones, both AMD models achieve higher average benchmark scores than supervised fine-tuning (SFT) baselines trained with the same sampling strategies. They also outperform the task-balancing methods evaluated in our experiments. Merging the two trained models can further improve their average benchmark score while yielding a single model for inference. The merged models outperform multi-task SFT by an average of 2.91 points across the three backbones.

\end{abstract}

\section{Introduction}

Supervised fine-tuning (SFT) on multi-task instruction mixtures is widely used
to adapt large language models (LLMs) to a range of tasks~
\citep{sanh2022multitask,chung2022scaling,wang-etal-2022-super}. These mixtures
often contain unequal amounts of data across tasks. Under proportional sampling,
tasks with more data are sampled more often. Increasing the emphasis on underrepresented tasks can improve their
performance, but may reduce performance on other tasks. Improving overall
multi-task performance requires managing these trade-offs.

Existing methods primarily manage these trade-offs by adjusting task contributions
within a single model. Data-mixture methods change how training examples are
sampled~\citep{xie2023doremi,wu-etal-2024-mixture-skills,shin2025dynamixsft},
while optimization-based methods reweight task losses or modify gradients~
\citep{chen2018gradnorm,yu2020pcgrad,liu2023famo}. Multi-task LLM fine-tuning
methods also regulate task contributions through convergence feedback,
task-specific masks, and task-balanced objectives~
\citep{gong-etal-2024-coba,liu2024mftcoder,zhang-etal-2025-mazo}.
Different strategies place different emphases on tasks and can produce models
with complementary strengths. These differences create an opportunity for
models to improve through collaboration.

Mutual distillation enables jointly trained models to exchange supervision~
\citep{zhang2018deep,anil2018large}. The value of this supervision depends on
the task and transfer direction. A model may provide useful guidance on one
task while benefiting from its partner's guidance on another. These
relationships can also change as both models learn. Distillation strength
should therefore adapt to the task, transfer direction, and stage of training.

\begin{figure}[t]
\centering
\includegraphics[width=\linewidth]{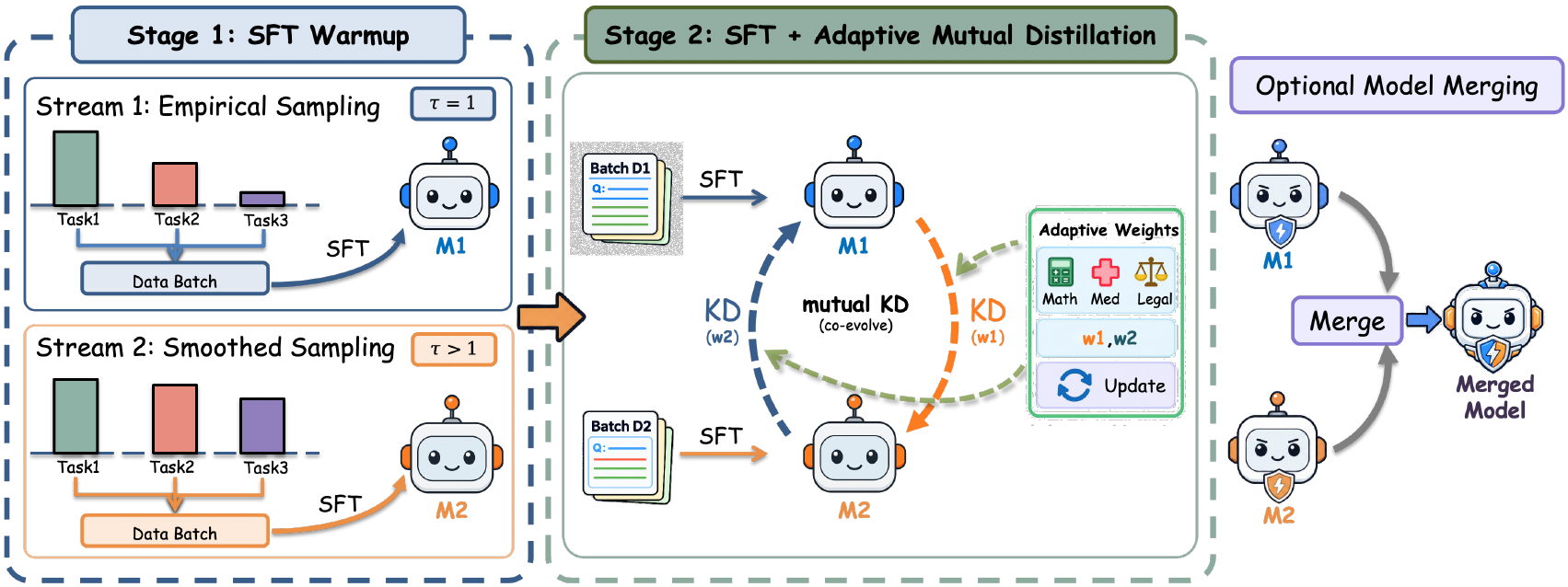}
\caption{Overview of \method{}.
Two models with different task-balancing strategies undergo independent SFT
warmup, followed by mutual distillation with task- and direction-specific
weights adjusted using validation feedback. The trained models can optionally
be merged into a single model for inference after collaborative training
concludes.}
\label{fig:method_overview}
\end{figure}

We propose \textbf{Adaptive Mutual Distillation (AMD)}, a
collaborative post-training framework that jointly trains two models with
different task-balancing strategies. In our experiments, one model samples tasks
in proportion to their frequencies in the training data. The other uses
temperature-smoothed sampling to increase the relative sampling probabilities
of tasks with less data.
Both models first undergo independent SFT warmup. They then continue training
under their respective sampling distributions while exchanging supervision
through mutual distillation. Task-wise validation feedback guides the
distillation weights for each task and transfer direction.
Figure~\ref{fig:method_overview} summarizes the two-stage framework and optional
post-training model merging.

At each weight update, \method{} runs three short training probes from the
current model pair, testing whether to decrease, maintain, or increase all
distillation weights. It then uses the resulting validation scores to select
an adjustment for each task and transfer direction.

We evaluate AMD on three LLM backbones and six benchmarks.
On each backbone, both AMD models achieve higher average benchmark scores than
SFT baselines trained with the same sampling strategies. They also outperform
the task-balancing methods evaluated in our experiments. Merging the two models
can further improve performance while yielding a single model for inference.
The merged models achieve the highest average benchmark scores among the
evaluated methods, exceeding multi-task SFT by
2.70--3.24 points.

Our contributions are as follows:
\begin{itemize}
  \item We propose AMD, a collaborative framework for multi-task LLM
  post-training with imbalanced task data. AMD jointly trains models with
  different task-balancing strategies and enables them to exchange task-wise
  supervision.
  \item We develop a validation-guided mechanism that evaluates candidate
  distillation-weight adjustments through shared training probes and selects
  updates separately for each task and transfer direction. It adapts supervision
  throughout training while keeping the number of training probes per update
  fixed as the number of tasks grows.
  \item Experiments on three LLM backbones and six benchmarks show that both
  AMD models outperform their sampling-matched SFT baselines in macro-average
  performance. Controlled comparisons support
  adapting distillation weights over training and separately for each task
  and transfer direction. Model merging provides additional gains while
  yielding a single model for inference.
\end{itemize}

\section{Related Work}

\paragraph{Multi-task learning.}
Task imbalance and interference are common challenges in multi-task learning.
Existing approaches adjust task losses, gradients, or the training data mixture
to regulate task contributions. GradNorm balances task training rates through
gradient normalization~\citep{chen2018gradnorm}, PCGrad modifies conflicting
gradients~\citep{yu2020pcgrad}, and FAMO adaptively balances task losses during
optimization~\citep{liu2023famo}. Optimization-based approaches also address
task balance in multi-task LLM fine-tuning. CoBa uses
convergence signals to adjust task weights~\citep{gong-etal-2024-coba},
MFTCoder uses balanced multi-task losses~\citep{liu2024mftcoder}, and MaZO
combines task-specific masks with zeroth-order
optimization~\citep{zhang-etal-2025-mazo}.
At the data level, DoReMi~\citep{xie2023doremi},
Mixture-of-Skills~\citep{wu-etal-2024-mixture-skills}, and
DynamixSFT~\citep{shin2025dynamixsft} control the relative contributions
of different data sources. HBO balances data allocation across and within
datasets~\citep{wang2026hbo}. \method{} brings models with different
task-balancing strategies into a collaborative training framework. Each model
retains its own strategy while learning from its partner through mutual
distillation.

\begin{figure}[t]
\centering
\includegraphics[width=\linewidth]{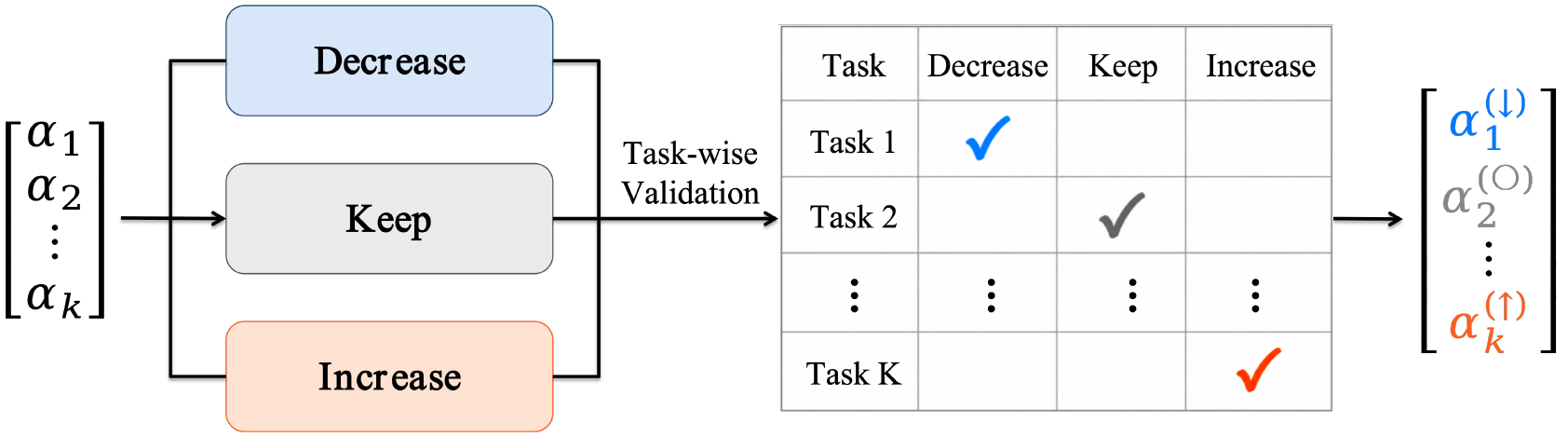}
\caption{Adaptive distillation-weight update.
Three probes globally decrease, retain, or increase the weights.
Validation scores determine the weight adjustment for each task and receiving
model.}
\label{fig:auto_amd_mechanism}
\end{figure}

\paragraph{Mutual distillation.}
Mutual distillation enables concurrently trained models to exchange
supervision~\citep{zhang2018deep,anil2018large}. OKDDip learns aggregation
weights to combine supervision from multiple peers~\citep{chen2020okddip}.
Online distillation has also been studied in multi-task learning, where jointly
trained single-task networks guide a multi-task network through adaptive feature
distillation and task weighting~\citep{jacob2023online}.
Other approaches adapt how teacher supervision is used. MMKD uses meta-learning
to adjust the contributions of multiple pretrained teachers based on feedback
from student updates~\citep{zhang2023mmkd}. DTO-KD dynamically balances task and
distillation losses using their gradients~\citep{hayder2026dtokd}.
\method{} jointly trains two multi-task models with different task-balancing
strategies. It uses task-wise validation scores from shared training probes to
adjust distillation weights separately for each task and transfer direction as
collaborative training progresses.

\section{Methodology}
\label{sec:method}

\method{} uses two training stages: independent SFT warmup and collaborative
training with adaptive mutual distillation. Validation feedback guides the
distillation weights for each task and transfer direction.
Algorithm~\ref{alg:adaptive_update} summarizes the adaptive weight updates used
in the second stage.
After training, the two models can optionally be merged into a single model
for inference (Section~\ref{sec:analysis_model_merging}).

\subsection{Framework}
\label{sec:method_framework}

We consider a training mixture in which each example has a task label. Both
models share the same data, architecture, and initialization. We index the tasks
participating in mutual distillation by $k\in\{1,\ldots,K\}$. General
instruction-following data are included in the training mixture for SFT, but do
not participate in mutual distillation.

\paragraph{Stage 1: SFT Warmup.}
During warmup, each model is trained independently with SFT. One model follows
Multi-task SFT (MFT), sampling tasks in proportion to their frequencies in the
training mixture. The other follows Smoothed SFT (SSFT), which raises these
frequencies to the power $1/\tau_s$, with $\tau_s>1$, and renormalizes them.
This increases the relative sampling probability of less frequent
tasks~\citep{arivazhagan2019massively,li-etal-2025-upsample}.

\paragraph{Stage 2: SFT with Adaptive Mutual Distillation.}
After warmup, each model continues training on batches drawn from its own
sampling distribution. For tasks participating in mutual distillation, its
partner provides token-level predictions on the same examples as distillation
targets. Both models are updated through their own training objectives, so the
supervision they provide evolves during training and reflects their current
token-level predictions.

Let $u\leftarrow v$ denote distillation from model $v$ to model $u$. For task
$k$, the receiving model uses the following training objective:
\begin{equation*}
\mathcal{L}
=
\bigl(1-\alpha_k^{u\leftarrow v}\bigr)\mathcal{L}_{\mathrm{SFT}}
+\alpha_k^{u\leftarrow v}\mathcal{L}_{\mathrm{KD}}.
\end{equation*}
Here, $\mathcal{L}_{\mathrm{SFT}}$ is the supervised fine-tuning loss.
$\mathcal{L}_{\mathrm{KD}}$ is the KL divergence from the supervising model's
token distribution to the receiving model's, averaged over response
tokens~\citep{hinton2015distilling}. The weight
$\alpha_k^{u\leftarrow v}\in[0,1]$ controls the distillation strength for this
task and direction.

\paragraph{Fixed-weight distillation.}
Under this objective, we consider two fixed-weight variants for comparison
with \method{}.
During Stage 2, \biregime{} uses the same fixed distillation weight
$\alpha_{\mathrm{Bi}}$ for all participating tasks and both transfer
directions, so each model receives supervision from its partner.
At each evaluation point in Stage 2, \uniregime{} selects the model
with the higher validation score on each participating task as the teacher
for that task. The teacher supervises its partner with the fixed weight
$\alpha_{\mathrm{Uni}}$ until the next evaluation.

\subsection{Adaptive-Weight Distillation}
\label{sec:adaptive_weight_update}

\method{} adjusts the distillation weight for each task and transfer direction
as training progresses. Shared training probes evaluate candidate adjustments,
and the receiving model's task-specific validation scores guide the selection
of each update.
Figure~\ref{fig:auto_amd_mechanism} illustrates this process
for one transfer direction.

\begin{algorithm}[t]
\caption{Adaptive update of task-wise directional weights}
\label{alg:adaptive_update}
\textbf{Input}: current model pair and directional weights
$\boldsymbol{\alpha}_s$\\
\textbf{Output}: updated directional weights $\boldsymbol{\alpha}_{s+1}$\par
\begin{algorithmic}[1]
\FOR{each action $a\in\{\downarrow,\circ,\uparrow\}$}
  \STATE Apply $a$ to all $2K$ weights to form
  $\boldsymbol{\alpha}_s^{(a)}$
  \STATE Start an independent probe from the same current model pair
  \STATE Run a short training probe using $\boldsymbol{\alpha}_s^{(a)}$
  \STATE Record validation scores $R_{k,u}^{(a)}$ for every task $k$ and model $u$
  \STATE Discard parameter changes made during the probe
\ENDFOR
\FOR{each task $k$ and direction $u\leftarrow v$}
  \STATE Choose $a_{k,u}^\star$ with the largest $R_{k,u}^{(a)}$
  \STATE Set $\alpha_{k,s+1}^{u\leftarrow v}$ to its value in
  $\boldsymbol{\alpha}_s^{(a_{k,u}^\star)}$
\ENDFOR
\STATE \textbf{return} $\boldsymbol{\alpha}_{s+1}$
\end{algorithmic}
\end{algorithm}

\paragraph{Global probing.}
At update round $s$, the vector $\boldsymbol{\alpha}_s$ contains the $2K$
task- and direction-specific distillation weights. \method{} constructs three
candidate configurations by decreasing, keeping, or increasing all entries of
this vector. These actions are denoted by
$a\in\{\downarrow,\circ,\uparrow\}$. Distillation weights are bounded mixing
coefficients. We therefore apply increases and decreases in logit space and
map the updated values back through the sigmoid function. This keeps the
weights within the valid range and yields smaller changes near the boundaries.
The logit-space step size is $\delta_s$; the keep action leaves the weights
unchanged.

Each candidate is evaluated through a short training probe starting from the
same current model pair. During each probe, both models train on shared batches
using the candidate weights (Appendix~\ref{app:hyperparameters}). Evaluation yields a validation
score $R_{k,u}^{(a)}$ for each task $k$ and model $u$. Parameter changes made
during the probe are then discarded.

\paragraph{Task-wise directional selection.}
For each task $k$ and direction $u\leftarrow v$, \method{} selects the action
with the highest validation score $R_{k,u}^{(a)}$ for the receiving model $u$.
The corresponding candidate weight becomes $\alpha_{k,s+1}^{u\leftarrow v}$.
The resulting vector $\boldsymbol{\alpha}_{s+1}$ can therefore combine weights
from different candidates. The original model pair resumes training with these
updated weights. Algorithm~\ref{alg:adaptive_update} summarizes the procedure
used at each adaptive update round.

\section{Experiments}
\label{sec:experiments}

We evaluate each AMD branch against its sampling-matched SFT baseline,
fixed-strength distillation variants, and prior task-balancing methods.
Tables~\ref{tab:main_results_qwen}--\ref{tab:main_results_llama} report these
comparisons across the evaluated backbones, together with the results of
post-training model merging.

\begin{table}[t]
\caption{Main results on Qwen3-0.6B-Base. Avg. is the unweighted mean across six
benchmarks. $\Delta$ Avg. is measured against Smoothed SFT for $\tau>1$ rows
and Multi-task SFT otherwise, including $+$ Merged. Bold indicates column-wise
best results.}
\label{tab:main_results_qwen}
  \centering

  \small
  \setlength{\tabcolsep}{1.5pt}
    \resizebox{\textwidth}{!}{%
    \begin{tabular}{llcccccccc}
      \toprule
      \multirow{2}{*}{\textbf{Method}} & \multirow{2}{*}{\textbf{Sampling}} &
      \multicolumn{2}{c}{\textbf{Math}} &
      \multicolumn{2}{c}{\textbf{Medicine}} &
      \multicolumn{2}{c}{\textbf{Law}} &
      \multirow{2}{*}{\textbf{Avg.}} &
      \multirow{2}{*}{\textbf{$\Delta$ Avg.}} \\
      \cmidrule(lr){3-4}\cmidrule(lr){5-6}\cmidrule(lr){7-8}
      & & \textbf{GSM8K} & \textbf{Math-500} &
      \textbf{MedQA} & \textbf{MedBullets} &
      \textbf{LawBench} & \textbf{CMMLU-Law} & & \\
      \midrule
      \multicolumn{10}{c}{\textbf{SFT Baselines}} \\
      Multi-task SFT & $\tau = 1$ & 57.47 & 42.40 & 27.34 & 40.10 & 29.53 & 48.99 & 40.97 & -- \\
      Smoothed SFT & $\tau > 1$ & 57.47 & 34.20 & 30.71 & 44.16 & 28.48 & 50.03 & 40.84 & -- \\
      \midrule
      \multicolumn{10}{c}{\textbf{Prior Methods}} \\
      PCGrad & $\tau = 1$ & 56.10 & 34.40 & 29.38 & 47.40 & 32.86 & 48.64 & 41.46 & +0.49 \\
      FAMO & $\tau = 1$ & 57.39 & 37.40 & 30.40 & 43.83 & 34.76 & 48.41 & 42.03 & +1.06 \\
      CoBa & $\tau = 1$ & 56.56 & 41.40 & 29.46 & 46.43 & 26.48 & 49.41 & 41.62 & +0.65 \\
      MFTCoder & $\tau = 1$ & 58.38 & 43.40 & 29.38 & 45.13 & 27.17 & 48.88 & 42.06 & +1.09 \\
      HBO & $\tau = 1$ & 60.42 & 33.60 & 29.54 & 42.05 & \textbf{34.93} & 49.58 & 41.69 & +0.72 \\
      \midrule
      \multicolumn{10}{c}{\textbf{Our Methods}} \\
      \multirow{2}{*}{Fixed-Bi} & $\tau = 1$ & 58.30 & 41.60 & 25.61 & 45.78 & 28.22 & 50.64 & 41.69 & +0.72 \\
      & $\tau > 1$ & 60.35 & 41.00 & 28.75 & 44.32 & 29.56 & 50.79 & 42.46 & +1.62 \\
      \noalign{\vskip 3pt}\hdashline\noalign{\vskip 3pt}
      \multirow{2}{*}{Fixed-Uni} & $\tau = 1$ & 60.27 & 43.20 & 24.98 & 43.18 & 26.88 & 50.20 & 41.45 & +0.48 \\
      & $\tau > 1$ & 57.39 & 41.40 & \textbf{33.94} & 48.70 & 27.96 & 49.76 & 43.19 & +2.35 \\
      \noalign{\vskip 3pt}\hdashline\noalign{\vskip 3pt}
      \multirow{2}{*}{Adaptive} & $\tau = 1$ & 60.05 & \textbf{44.00} & 30.40 & 44.97 & 31.40 & 49.85 & 43.45 & +2.48 \\
      & $\tau > 1$ & 60.65 & 40.80 & 28.91 & 46.43 & 31.66 & 50.67 & 43.19 & +2.35 \\
      \rowcolor[gray]{0.94}
      \multicolumn{2}{l}{\hspace{0.8em}$+$ Merged} & \textbf{61.11} & 42.60 & 28.67 & \textbf{49.03} & 29.90 & \textbf{51.17} & \textbf{43.75} & \textbf{+2.78} \\
      \bottomrule
    \end{tabular}%
    }
\end{table}

\subsection{Setup}
\label{sec:settings}

\paragraph{Experimental setting.}
We evaluated \method{} on Qwen3-0.6B-Base~\citep{yang2025qwen3},
Llama-3.1-8B~\citep{grattafiori2024llama}, and
Llama-3.2-1B~\citep{meta2024llama32modelcard}. The training mixture covered
mathematics, medicine, law, and general instruction following. It contained
100k examples from NuminaMath~\citep{aimo2025numinamath15}, 9,178 from
MedQA~\citep{jin2021medqa}, 50k from
DISC-Law~\citep{yue2023disclawllm}, and 10k from
InfinityInstruct~\citep{li2025infinityinstruct}.

The two AMD models shared this mixture. One used empirical sampling
($\tau=1$), and the other used temperature-smoothed sampling ($\tau_s=3$).
We used full-parameter SFT with a learning rate of $1\times10^{-5}$ for
Qwen3-0.6B-Base and Llama-3.2-1B. For Llama-3.1-8B, we used LoRA with rank 16,
alpha 32, dropout 0.05, and the target modules set to \texttt{all}.
The learning rate was $5\times10^{-5}$.
In the main experiments, all models were trained for three
epochs with an effective batch size of 64. Mutual distillation began at step
3000 with temperature $T=1.0$.

\paragraph{Distillation configurations.}
Across all three backbones, Fixed-Bi and Fixed-Uni used fixed distillation weights of
0.4 and 0.2, respectively. Adaptive initialized all task- and direction-specific
weights to 0.4 and updated them within $[0,1]$. Adaptive used a
logit-space step size of $\delta_s=\mu_s\delta_0$, with $\delta_0=0.5$.
We set $\mu_s=\sqrt{1-t/T_{\max}}$, where $t/T_{\max}$ is the fraction of
training completed. After warmup, we updated the distillation weights every
1000 training steps.
Fixed-Uni used the same interval for teacher selection. Each Adaptive probe
used 1\% of the one-epoch training budget. Adaptive updates used task-wise
validation accuracy.

\paragraph{Evaluation.}
\begin{table*}[t]
    \caption{Main results on Llama-3.1-8B.}
    \label{tab:main_results_llama31_8b}
  \centering

  \small
  \setlength{\tabcolsep}{1.5pt}
    \resizebox{\textwidth}{!}{%
    \begin{tabular}{llcccccccc}
      \toprule
      \multirow{2}{*}{\textbf{Method}} & \multirow{2}{*}{\textbf{Sampling}} &
      \multicolumn{2}{c}{\textbf{Math}} &
      \multicolumn{2}{c}{\textbf{Medicine}} &
      \multicolumn{2}{c}{\textbf{Law}} &
      \multirow{2}{*}{\textbf{Avg.}} &
      \multirow{2}{*}{\textbf{$\Delta$ Avg.}} \\
      \cmidrule(lr){3-4}\cmidrule(lr){5-6}\cmidrule(lr){7-8}
      & & \textbf{GSM8K} & \textbf{Math-500} &
      \textbf{MedQA} & \textbf{MedBullets} &
      \textbf{LawBench} & \textbf{CMMLU-Law} & & \\
      \midrule
      \multicolumn{10}{c}{\textbf{SFT Baselines}} \\
      Multi-task SFT & $\tau = 1$ & 58.45 & 27.20 & 49.25 & 51.14 & 28.46 & \textbf{50.49} & 44.16 & -- \\
      Smoothed SFT & $\tau > 1$ & 50.42 & 26.00 & 50.67 & 55.19 & 35.84 & 47.99 & 44.35 & -- \\
      \midrule
      \multicolumn{10}{c}{\textbf{Prior Methods}} \\
      PCGrad & $\tau = 1$ & 62.32 & 25.80 & 50.12 & 55.36 & 31.93 & 49.92 & 45.91 & +1.75 \\
      FAMO & $\tau = 1$ & 62.17 & 22.80 & 50.59 & 54.71 & 31.25 & 47.53 & 44.84 & +0.68 \\
      CoBa & $\tau = 1$ & 61.87 & 29.00 & 51.69 & 51.46 & 20.35 & 48.51 & 43.81 & -0.35 \\
      MFTCoder & $\tau = 1$ & 63.61 & 26.20 & \textbf{52.08} & 52.11 & 28.40 & 48.96 & 45.23 & +1.07 \\
      HBO & $\tau = 1$ & \textbf{66.57} & 21.00 & \textbf{52.08} & \textbf{56.66} & 33.08 & 45.01 & 45.73 & +1.57 \\
      \midrule
      \multicolumn{10}{c}{\textbf{Our Methods}} \\
      \multirow{2}{*}{Fixed-Bi} & $\tau = 1$ & 61.49 & 26.80 & 50.59 & 53.73 & 28.82 & 50.22 & 45.27 & +1.11 \\
      & $\tau > 1$ & 62.93 & 27.80 & 48.86 & 56.01 & 33.46 & 47.60 & 46.11 & +1.76 \\
      \noalign{\vskip 3pt}\hdashline\noalign{\vskip 3pt}
      \multirow{2}{*}{Fixed-Uni} & $\tau = 1$ & 63.00 & 24.20 & 51.37 & 53.90 & 33.32 & 47.76 & 45.59 & +1.43 \\
      & $\tau > 1$ & 64.06 & 29.60 & 49.96 & 55.03 & 32.89 & 47.10 & 46.44 & +2.09 \\
      \noalign{\vskip 3pt}\hdashline\noalign{\vskip 3pt}
      \multirow{2}{*}{Adaptive} & $\tau = 1$ & 63.61 & 27.60 & 51.69 & 52.60 & 34.17 & 50.30 & 46.66 & +2.50 \\
      & $\tau > 1$ & 63.23 & 29.40 & 49.65 & 54.87 & \textbf{36.41} & 47.79 & 46.89 & +2.54 \\
      \rowcolor[gray]{0.94}
      \multicolumn{2}{l}{\hspace{0.8em}$+$ Merged} & 65.13 & \textbf{29.80} & 51.37 & 55.52 & 33.68 & 48.92 & \textbf{47.40} & \textbf{+3.24} \\
      \bottomrule
    \end{tabular}%
    }
\end{table*}

\begin{table*}[t]
\caption{Main results on Llama-3.2-1B.}
\label{tab:main_results_llama}
  \centering
  \small
  \setlength{\tabcolsep}{1.5pt}
    \resizebox{\textwidth}{!}{%
    \begin{tabular}{llcccccccc}
      \toprule
      \multirow{2}{*}{\textbf{Method}} & \multirow{2}{*}{\textbf{Sampling}} &
      \multicolumn{2}{c}{\textbf{Math}} &
      \multicolumn{2}{c}{\textbf{Medicine}} &
      \multicolumn{2}{c}{\textbf{Law}} &
      \multirow{2}{*}{\textbf{Avg.}} &
      \multirow{2}{*}{\textbf{$\Delta$ Avg.}} \\
      \cmidrule(lr){3-4}\cmidrule(lr){5-6}\cmidrule(lr){7-8}
      & & \textbf{GSM8K} & \textbf{Math-500} &
      \textbf{MedQA} & \textbf{MedBullets} &
      \textbf{LawBench} & \textbf{CMMLU-Law} & & \\
      \midrule
      \multicolumn{10}{c}{\textbf{SFT Baselines}} \\
      Multi-task SFT & $\tau = 1$ & 19.94 & 7.00 & 30.40 & 29.55 & 16.55 & 33.53 & 22.83 & -- \\
      Smoothed SFT & $\tau > 1$ & 20.39 & 6.80 & 28.20 & 32.79 & 16.70 & 32.94 & 22.97 & -- \\
      \midrule
      \multicolumn{10}{c}{\textbf{Prior Methods}} \\
      PCGrad & $\tau = 1$ & 15.62 & 5.80 & 27.65 & 35.88 & 22.15 & 32.20 & 23.22 & +0.39 \\
      FAMO & $\tau = 1$ & 21.99 & 6.80 & 30.95 & 29.55 & 21.53 & 30.43 & 23.54 & +0.71 \\
      CoBa & $\tau = 1$ & 18.42 & 8.20 & 27.42 & 33.12 & 21.03 & 31.93 & 23.35 & +0.52 \\
      MFTCoder & $\tau = 1$ & 19.33 & 9.00 & 26.00 & 34.90 & 20.86 & 30.81 & 23.48 & +0.65 \\
      HBO & $\tau = 1$ & 17.51 & 6.20 & 29.22 & \textbf{36.36} & 20.25 & 34.04 & 23.93 & +1.10 \\
      \midrule
      \multicolumn{10}{c}{\textbf{Our Methods}} \\
      \multirow{2}{*}{Fixed-Bi} & $\tau = 1$ & 23.58 & 7.60 & 31.50 & 33.44 & 16.41 & 31.90 & 24.07 & +1.24 \\
      & $\tau > 1$ & 23.65 & 8.80 & \textbf{32.05} & 30.19 & 17.25 & 31.57 & 23.92 & +0.95 \\
      \noalign{\vskip 3pt}\hdashline\noalign{\vskip 3pt}
      \multirow{2}{*}{Fixed-Uni} & $\tau = 1$ & 21.91 & 7.60 & 29.38 & 32.31 & 21.29 & 33.79 & 24.38 & +1.55 \\
      & $\tau > 1$ & 20.55 & \textbf{9.20} & 29.77 & 33.60 & \textbf{23.82} & 30.11 & 24.51 & +1.54 \\
      \noalign{\vskip 3pt}\hdashline\noalign{\vskip 3pt}
      \multirow{2}{*}{Adaptive} & $\tau = 1$ & \textbf{25.32} & 7.60 & 28.83 & 34.58 & 20.93 & 32.92 & 25.03 & +2.20 \\
      & $\tau > 1$ & 22.44 & 7.80 & 26.87 & 34.09 & 21.76 & \textbf{34.29} & 24.54 & +1.57 \\
      \rowcolor[gray]{0.94}
      \multicolumn{2}{l}{\hspace{0.8em}$+$ Merged} & 24.56 & 7.20 & 31.89 & 35.88 & 20.69 & 32.98 & \textbf{25.53} & \textbf{+2.70} \\
      \bottomrule
    \end{tabular}%
    }
\end{table*}

The evaluation covered six benchmarks across three domains:
GSM8K~\citep{cobbe2021gsm8k} and
Math-500~\citep{hendrycks2021math,lightman2024lets} for mathematics,
MedQA~\citep{jin2021medqa} and MedBullets~\citep{chen-etal-2025-benchmarking}
for medicine, and LawBench~\citep{fei-etal-2024-lawbench} and
CMMLU-Law~\citep{li-etal-2024-cmmlu} for law. All evaluations used OpenCompass
v0.5.3~\citep{2023opencompass}. Overall multi-task performance was measured by
the macro-average score, defined as the unweighted mean of the six benchmark
scores. Checkpoints were selected using validation performance; the validation
and model-selection protocols are detailed in
Appendix~\ref{app:validation_selection}.

\paragraph{Comparisons.}
We compared \method{} with MFT and SSFT baselines trained independently using
the same sampling distributions as the corresponding AMD models. We also
evaluated prior single-model task-balancing methods: PCGrad~\citep{yu2020pcgrad},
FAMO~\citep{liu2023famo}, CoBa~\citep{gong-etal-2024-coba},
MFTCoder~\citep{liu2024mftcoder}, and HBO~\citep{wang2026hbo}. These methods
began from the empirical task mixture. Fixed-Bi and Fixed-Uni provided
fixed-strength comparisons for adaptive distillation. Within each backbone,
all methods used the same training examples, tokenizer, generation settings,
and evaluation protocol.

\subsection{Main Results}
\label{sec:main_results}

Tables~\ref{tab:main_results_qwen}--\ref{tab:main_results_llama} compare AMD
with SFT baselines and prior task-balancing methods across three LLM backbones.
On all three backbones, both Adaptive branches achieve higher macro-average
scores than their corresponding SFT baselines. They also outperform all
evaluated prior task-balancing methods. Relative to the corresponding SFT baselines, the gains
range from 2.20 to 2.50 points for the empirical-sampling branch and from 1.57
to 2.54 points for the smoothed-sampling branch.

Among the distillation variants, Adaptive produces the individual model with
the highest macro-average score on each backbone. Its highest-scoring branch exceeds the best
fixed-distillation branch by 0.26, 0.45, and 0.52 points on Qwen3-0.6B-Base,
Llama-3.1-8B, and Llama-3.2-1B, respectively. The gains vary across sampling
branches: on Qwen3-0.6B-Base, Adaptive and Fixed-Uni tie under smoothed sampling.

Merging the two Adaptive models further improves macro-average performance. The merged
models achieve the highest macro-average scores in all three main tables,
reaching 43.75, 47.40, and 25.53 on Qwen3-0.6B-Base, Llama-3.1-8B, and
Llama-3.2-1B, respectively. These scores exceed those of the better Adaptive
branch by 0.30, 0.51, and 0.50 points, respectively. Across the three backbones,
the merged models improve over multi-task SFT by an average of 2.91 points.

\section{Analysis}
\label{sec:analysis}

We analyze \method{} on Llama-3.2-1B, starting with sampling configurations
and adaptive distillation. We then examine variability across seeds and the
additional gains from model merging.
\mbox{Appendices~\ref{app:fixed_weight_sweep}--\ref{app:longer_budget_stress}}
examine fixed distillation weights, weight-update schedules, and training
budgets.

\subsection{Effect of Sampling Distributions}
\label{sec:analysis_diverse_sampling}

\begin{table}[t]
  \caption{Component comparisons on Llama-3.2-1B. Scores are six-benchmark
  macro-averages. Sampling lists the two temperatures; paired scores follow
  the same model order. Task-Tied MD selects one action
  per direction, shared across tasks.}
  \label{tab:component_comparison}
  \label{tab:same_temp_seed_control}
  \label{tab:pmd_variant_comparison}
  \centering
  \small
  \setlength{\tabcolsep}{3pt}
  \renewcommand{\arraystretch}{1.0}
  \begin{minipage}[t]{0.48\linewidth}
    \vspace{0pt}
    \centering
    \begin{tabular}[b]{@{}p{0.44\linewidth}@{}>{\centering\arraybackslash}p{0.22\linewidth}@{}>{\raggedleft\arraybackslash}p{0.34\linewidth}@{}}
      \toprule
      \textbf{Method} & \textbf{Sampling} & \textbf{Avg.} \\
      \midrule
      SFT baselines & $(1,3)$ & 22.83 / 22.97 \\
      \addlinespace[2pt]
      Same-temp MD & $(1,1)$ & 22.50 / 22.45 \\
      \addlinespace[2pt]
      Same-temp MD & $(3,3)$ & 23.18 / 22.22 \\
      \addlinespace[2pt]
      Fixed-Bi & $(1,3)$ & 24.07 / 23.92 \\
      \midrule
      Adaptive & $(1,3)$ & \textbf{25.03} / \textbf{24.54} \\
      \bottomrule
    \end{tabular}
    \par\vspace{0pt}
  \end{minipage}\hfill
  \begin{minipage}[t]{0.48\linewidth}
    \vspace{0pt}
    \centering
\begin{tabular}[b]{@{}p{0.44\linewidth}@{}>{\centering\arraybackslash}p{0.22\linewidth}@{}>{\raggedleft\arraybackslash}p{0.34\linewidth}@{}}
  \toprule
  \textbf{Method} & \textbf{Sampling} & \textbf{Avg.} \\
  \midrule
  Globally Tied MD & $(3,3)$ & 23.05 / 23.41 \\
  \addlinespace[2pt]
  Final-Weight MD & $(1,3)$ & 23.41 / 22.01 \\
  \addlinespace[2pt]
  Direction-Tied MD & $(1,3)$ & 24.59 / 23.38 \\
  \addlinespace[2pt]
  Task-Tied MD & $(1,3)$ & 24.33 / 23.69 \\
  \midrule
  Adaptive & $(1,3)$ & \textbf{25.03} / \textbf{24.54} \\
  \bottomrule
\end{tabular}

    \par\vspace{0pt}
  \end{minipage}
  \par\vspace{3pt}
  \begin{minipage}[t]{0.48\linewidth}
    \centering
    \textbf{(a) Sampling configurations}
  \end{minipage}\hfill
  \begin{minipage}[t]{0.48\linewidth}
    \centering
    \textbf{(b) Distillation-weight policies}
  \end{minipage}
\end{table}

Table~\ref{tab:same_temp_seed_control}(a) compares mutual distillation when
the two models use the same or different sampling distributions.
Same-temp MD uses the same fixed
bidirectional distillation weights as \biregime{}, with different sampling
seeds for the two models. Under empirical sampling, the two models achieve
22.50 and 22.45, both below the corresponding SFT baseline of 22.83. Under
smoothed sampling, they achieve 23.18 and 22.22, with one above and the other
below the corresponding SFT baseline of 22.97. With different sampling
distributions, \biregime{} reaches 24.07 and 23.92, outperforming both
corresponding SFT baselines. Its highest branch score exceeds the best
same-temperature result by 0.89 points. These results support using different
sampling distributions for the two models under fixed bidirectional distillation.

\subsection{Effect of Adaptive Weight Updates}
\label{sec:analysis_controlled_md_variants}

Online adaptation yields higher scores in both branches than keeping the
learned final weights fixed throughout collaborative training.
Following the same SFT warmup, Final-Weight MD keeps Adaptive's final task-
and direction-specific distillation weights fixed throughout Stage 2.
It achieves 23.41 and 22.01
under empirical and smoothed sampling, respectively, compared with 25.03 and
24.54 for Adaptive (Table~\ref{tab:pmd_variant_comparison}(b)).

Updating distillation weights separately for each transfer direction yields
higher scores in both branches than Direction-Tied MD.
Direction-Tied MD retains task-specific
adaptation but ties the two directional weights within each task. At each
update, it averages the two branches' task-specific validation scores for each
candidate action to select a shared action for both directions. It achieves
24.59 under empirical sampling and 23.38 under smoothed sampling
(Table~\ref{tab:pmd_variant_comparison}(b)).

Updating distillation weights separately for each task yields higher scores
in both branches than Task-Tied MD.
Task-Tied MD retains direction-specific adaptation but selects a single
update action per direction and applies it to all tasks. It achieves 24.33
under empirical sampling and 23.69 under smoothed sampling
(Table~\ref{tab:pmd_variant_comparison}(b)).

To assess whether a shared update action benefits both models under matched
smoothed sampling, we compare Globally Tied MD with fixed-weight Same-temp MD.
Globally Tied MD selects one action across tasks and transfer directions using
a weighted average of task-wise validation scores. Its best branch reaches
23.41, compared with the best Same-temp MD branch at 23.18
(Table~\ref{tab:component_comparison}(a,b)). The shared update improves one
branch but slightly lowers the other relative to Same-temp MD.

\begin{figure}[t]
  \centering
  \includegraphics[width=\linewidth]{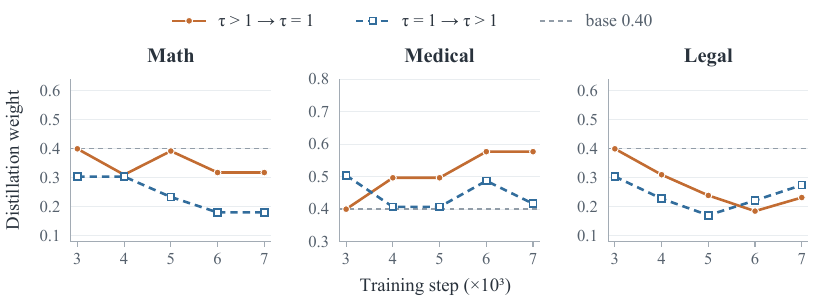}
  \caption{Task- and direction-specific distillation-weight trajectories on
  Llama-3.2-1B. The first point in each trajectory shows the weight after the
  initial adaptive update at step 3000.}
  \label{fig:auto_pmd_weight_trajectories}
\end{figure}

To inspect how task- and direction-specific weights evolve during training,
Figure~\ref{fig:auto_pmd_weight_trajectories} shows trajectories for Math,
Medicine, and Law. In Math, both weights decrease overall, with a generally
higher weight for the smoothed-to-empirical direction. In Medicine, this
direction increases while the reverse direction fluctuates, changing their
relative ordering. In Law, both weights decrease initially and then rise
toward the end of training. These different trajectories illustrate how
Adaptive changes distillation strength by task and transfer direction.

\subsection{Variability Across Training Seeds}
\label{sec:analysis_training_seeds}

\begin{table}[htbp]
  \caption{Training-seed variability of the three \method{} regimes on
  Llama-3.2-1B. Scores are unweighted six-benchmark macro-averages, summarized
  across three runs as mean $\pm$ sample standard deviation. The $+$ Merged
  row reports the merged Adaptive model from each run.}
  \label{tab:training_seed_robustness}
  \centering
  \small
  \setlength{\tabcolsep}{7pt}
  \renewcommand{\arraystretch}{1.08}
  \begin{tabular}{@{}llcccc@{}}
    \toprule
    \multirow{2}{*}{\textbf{Method}} &
    \multirow{2}{*}{\textbf{Sampling}} &
    \multicolumn{3}{c}{\textbf{Macro Avg. by seed}} &
    \multirow{2}{*}{\textbf{Mean $\pm$ Std.}} \\
    \cmidrule(lr){3-5}
    & & \makebox[0.9cm]{42} & \makebox[0.9cm]{43} &
    \makebox[0.9cm]{44} & \\
    \midrule
    \multirow{2}{*}{\biregime{}}
      & $\tau=1$ & 24.07 & 23.79 & 23.83 & 23.90 $\pm$ 0.15 \\
      & $\tau=3$ & 23.92 & 24.14 & 24.44 & 24.17 $\pm$ 0.26 \\
    \addlinespace[3pt]
    \multirow{2}{*}{\uniregime{}}
      & $\tau=1$ & 24.38 & 24.52 & 24.16 & 24.35 $\pm$ 0.18 \\
      & $\tau=3$ & 24.51 & 24.40 & 24.45 & 24.45 $\pm$ 0.06 \\
    \addlinespace[3pt]
    \multirow{2}{*}{\autoregime{}}
      & $\tau=1$ & 25.03 & 24.32 & 24.44 & 24.60 $\pm$ 0.38 \\
      & $\tau=3$ & 24.54 & 24.70 & 24.52 & 24.59 $\pm$ 0.10 \\
    \rowcolor[gray]{0.94}
    \multicolumn{2}{l}{\hspace{0.8em}$+$ Merged}
      & 25.53 & 25.61 & 25.28 & 25.47 $\pm$ 0.17 \\
    \bottomrule
  \end{tabular}
\end{table}

To assess whether the gains from adaptive distillation persist across training
seeds, we repeat each distillation regime with two additional seeds, yielding
three runs per regime (Table~\ref{tab:training_seed_robustness}).

Adaptive achieves 24.60 $\pm$ 0.38 under empirical sampling and
24.59 $\pm$ 0.10 under smoothed sampling. Both branches outperform their
corresponding Fixed-Bi and Fixed-Uni baselines on average. The
smoothed-sampling branch also exceeds both corresponding baselines in every
reported run. The averaged gains for both branches and the per-run gains under
smoothed sampling support the repeatability of \method{}'s performance
improvements across training seeds.

\subsection{Post-training Model Merging}
\label{sec:analysis_model_merging}

We evaluate post-training model merging as a way to obtain additional gains
from the two \method{} models while using a single model for inference.
Using MergeKit~\citep{goddard-etal-2024-arcees}, we average the parameters of
the two models from the same training step with equal weights. We apply the
same procedure to the independently trained Multi-task SFT and Smoothed SFT
models.

\begin{table}[htbp]
  \caption{Model merging on Llama-3.2-1B. Scores are six-benchmark
  macro-averages; $\Delta$ Merge is the gain over the better model before merging.}
  \label{tab:model_merging_comparison}
  \centering
  \small
  \setlength{\tabcolsep}{7pt}
  \renewcommand{\arraystretch}{1.08}
  \begin{tabular}{@{}lcccc@{}}
    \toprule
    \textbf{Training} & \shortstack{\textbf{Empirical}\\\textbf{sampling}} &
    \shortstack{\textbf{Smoothed}\\\textbf{sampling}} &
    \textbf{Merged} & \textbf{$\Delta$ Merge} \\
    \midrule
    SFT baselines & 22.83 & 22.97 & 24.08 & +1.11 \\
    \method{} & 25.03 & 24.54 & \textbf{25.53} & +0.50 \\
    \bottomrule
  \end{tabular}
\end{table}

The merged SFT baseline provides a reference for assessing the benefit of
collaborative training beyond averaging independently trained models.
Both \method{} branches outperform this baseline, which scores 24.08
(Table~\ref{tab:model_merging_comparison}). Their advantage is therefore
already present before merging, supporting the contribution of collaborative
training to the performance of the constituent models.

Merging further raises \method{}'s score from 25.03 for its better branch
to 25.53, exceeding the merged SFT baseline by 1.45 points.
Across three training seeds, the merged \method{} model achieves
25.47 $\pm$ 0.17 and outperforms both constituent models in every run
(Table~\ref{tab:training_seed_robustness}). Parameter averaging thus builds
on the gains from collaborative training and combines the two models into
a stronger model for inference with the same backbone architecture.

\section{Conclusion}

\method{} addresses imbalanced multi-task LLM post-training by enabling models
with different sampling distributions to learn from each other. Validation
feedback guides the distillation strength for each task and transfer direction
throughout training. Across three LLM backbones and six benchmarks, both trained
models outperform their sampling-matched SFT baselines in macro-average
performance. Optional parameter merging provides further gains and produces a
single model for inference. The merged models achieve the highest macro-average
scores among the compared methods, exceeding multi-task SFT by 2.70--3.24 points.
These results support adaptive collaboration between models with different task
emphases for improving overall multi-task performance.

\newpage
\section*{AI Use Statement}
Generative AI tools were used only to polish the language of the manuscript.
They were not used to assist with translation or to interpret results.
The authors take responsibility for the final content.

\section*{Reproducibility Statement}
The training framework and adaptive weight updates are described in
Section~\ref{sec:method} and Algorithm~\ref{alg:adaptive_update}.
Section~\ref{sec:settings} and Appendix~\ref{app:experimental_details}
document the datasets, backbone models, training settings, evaluation protocol,
and checkpoint-selection procedure. Section~\ref{sec:analysis_training_seeds}
reports variability across training seeds.

\bibliography{custom}
\bibliographystyle{iclr2027_conference}

\clearpage
\appendix
\section*{Appendix}
\noindent
The appendix provides experimental details and additional analyses of
temperature-smoothed baselines, fixed distillation weights, weight-update
schedules, and training budgets.

\section{Experimental Protocol}
\label{app:experimental_details}

\subsection{Data Mixture and Evaluation}
\label{app:dataset_details}

The training mixture assigns one fixed task category to each example before
training. Table~\ref{tab:appendix_training_mixture} summarizes the
category-level datasets and sample caps used in the main experiments. These
task categories are used consistently by the sampler, the single-model
multi-task baselines, and the \method{} distillation policies.

\begin{table}[ht]
  \caption{Training mixture for the main experiments. Each row lists one
  task category, its source dataset, and the configured sample cap.}
  \label{tab:appendix_training_mixture}
  \centering
  \small
  \setlength{\tabcolsep}{2pt}
  \begin{tabular}{llr}
    \toprule
    \textbf{Category} & \textbf{Dataset} & \textbf{Cap} \\
    \midrule
    Math & NuminaMath-1.5~\citep{aimo2025numinamath15} & 100k \\
    Medicine & MedQA~\citep{jin2021medqa} & 9,178 \\
    Law & DISC-Law~\citep{yue2023disclawllm} & 50k \\
    General & InfinityInstruct~\citep{li2025infinityinstruct} & 10k \\
    \bottomrule
  \end{tabular}
\end{table}

For evaluation, all results reported within a backbone family use the same six
benchmarks and post-processing rules. We use two benchmarks for each of the
mathematical reasoning, medical reasoning, and legal reasoning categories. The
reported aggregate score is the macro-average over these six benchmark scores.
LawBench evaluation covered the full benchmark, and CMMLU-Law included all
law-related subsets of CMMLU.
Training, validation, and test data were mutually disjoint.

\paragraph{Validation and model selection.}
\label{app:validation_selection}
Adaptive updates used accuracy on validation subsets from the validation sets
provided with MATH, MedQA, and LawBench for the mathematics, medicine, and law
tasks, respectively. We randomly sampled 100 validation examples for
mathematics, 100 for medicine, and 200 for law. These subsets remained fixed
throughout training and were shared across all three backbones.
For each task and receiving model, we use the accuracy returned by OpenCompass
on the corresponding fixed validation subset as $R_{k,u}^{(a)}$ to compare
the candidate probes.
General instruction-following data
were used for SFT only and did not contribute to adaptive validation feedback.
For \method{}, the checkpoint was selected by taking the unweighted mean of
the six validation accuracies across both branches and the three tasks.
The training step with the highest mean was selected, and both models at that
step were used for branch-wise reporting. The same pair was then merged to
produce the $+$ Merged result, without a separate search over merged
candidates or further training. Single-model methods used the checkpoint with
the highest mean validation score.

\subsection{Training Configuration}
\label{app:hyperparameters}

Unless a method requires a different batch structure, we keep the data mixture,
tokenizer, cutoff length, learning-rate schedule, task-utility mapping, and
six-benchmark evaluation protocol fixed within each backbone family. Training
runs are conducted on 8$\times$A800-SXM4-80GB GPUs. The main Llama-3.2-1B and
Qwen3-0.6B-Base runs use full-parameter SFT with sequence cutoff length 2048,
bf16 training, cosine scheduling, warmup ratio 0.1, and learning rate
$1\mathrm{e}{-5}$. The Llama-3.1-8B runs use LoRA rather than full-parameter
updates due to the larger backbone size.

The two-model \method{} setup uses two task-sampling temperatures: $\tau=1$
preserves the empirical task mixture, while $\tau>1$ increases the relative
sampling probability of smaller task categories. In the reported main runs, the
two models use $\tau=1$ and $\tau=3$. \method{} enables cross-model
distillation after the initial SFT phase; the main policy runs start
distillation at step 3000 and use distillation temperature $T=1.0$.
Method-specific policy settings are fixed before test evaluation and are read
from the corresponding training configuration rather than tuned per reported
benchmark.

Each probe uses the next batches that would be sampled along the current
training trajectory. Both models share these batches, and all candidates use
the same batch sequence. Model, optimizer, learning-rate scheduler, data
iterator, and random number generator states are restored to their pre-probe
values before each candidate and before training resumes with the selected
weights. When validation scores tie, Adaptive leaves the corresponding
distillation weight unchanged, while Fixed-Uni retains the current teacher
direction.

To keep the effective optimizer-update scale comparable, standard two-model
runs use two models with per-device batch size 4 and gradient accumulation 2.
MFTCoder and CoBa use per-device batch size 8 and gradient accumulation 1,
while PCGrad and FAMO use task-grouped micro-batches so that each optimizer
update covers all four task categories. This keeps the comparison data-matched,
although not exactly wall-clock-cost matched.

\subsection{Rationale for Shared Probing and Task-wise Selection}
\label{app:shared_probing_rationale}

\method{} uses shared training probes to obtain task-specific feedback on
candidate distillation-weight adjustments. Each probe starts from the same
current model and optimizer states and applies an adjustment to the full weight
vector. Its validation results measure each receiving model's performance on
each task after training under that candidate configuration. Selecting a single
action from an aggregate score forces all tasks and transfer directions to share
the same adjustment, even when their validation scores favor different
candidates. \method{} preserves these differences by using each task's
validation ranking to select the corresponding directional weight adjustment.

The selected adjustments are combined into a new weight vector, and training
resumes from the original state with these weights. This feedback-guided
recombination enables task- and direction-specific updates while keeping the
number of training probes per update fixed as the number of tasks grows.
Because the combined vector can differ from every probed configuration, its
performance need not match the individual probe scores. The probe results
provide guidance for constructing the update, while the resulting training
performance determines its effectiveness.

For each transfer direction, Task-Tied MD computes the unweighted mean
validation score across tasks for each candidate action. It selects the action
with the highest mean score and applies it to all tasks in that direction.

\section{Temperature-Smoothed Baseline Setting}
\label{app:prior_baselines_tau_gt1}

Table~\ref{tab:prior_baselines_tau_gt1} reports a sampling-matched
$\tau>1$ control on Llama-3.2-1B. The table includes the temperature-smoothed
SFT reference, prior single-model baselines, and the corresponding
smoothed-sampling branches of the mutual-distillation regimes. Each reported
model is trained with the same temperature-smoothed sampling distribution.
The table uses the same six benchmark columns and
macro-average definition as Table~\mainref{tab:main_results_llama}.

\newpage
\begin{table}[H]
\caption{Sampling-matched comparison on Llama-3.2-1B under
temperature-smoothed sampling. Each reported model or branch uses the same
$\tau>1$ sampling distribution and six-benchmark evaluation protocol. All rows report raw scores;
$\Delta$ Avg. gives the macro-average gain relative to the Smoothed
SFT reference. Bold marks the highest value in each column.}
  \label{tab:prior_baselines_tau_gt1}
  \centering

  \small
  \setlength{\tabcolsep}{1.5pt}
  \renewcommand{\arraystretch}{1.06}
    \resizebox{\textwidth}{!}{%
    \begin{tabular}{llcccccccc}
      \toprule
      \multirow{2}{*}{\textbf{Method}} &
      \multirow{2}{*}{\textbf{Sampling}} &
      \multicolumn{2}{c}{\textbf{Math}} &
      \multicolumn{2}{c}{\textbf{Medicine}} &
      \multicolumn{2}{c}{\textbf{Law}} &
      \multirow{2}{*}{\textbf{Avg.}} &
      \multirow{2}{*}{\textbf{$\Delta$ Avg.}} \\
      \cmidrule(lr){3-4}\cmidrule(lr){5-6}\cmidrule(lr){7-8}
      & & \textbf{GSM8K} & \textbf{Math-500} &
      \textbf{MedQA} & \textbf{MedBullets} &
      \textbf{LawBench} & \textbf{CMMLU-Law} & & \\
      \midrule
      \addlinespace[2pt]
      \multicolumn{10}{c}{\textbf{SFT Baselines}} \\
      \addlinespace[1pt]
      Smoothed SFT & $\tau > 1$ & 20.39 & 6.80 & 28.20 & 32.79 & 16.70 & 32.94 & 22.97 & -- \\
      \midrule
      \addlinespace[2pt]
      \multicolumn{10}{c}{\textbf{Prior Methods}} \\
      \addlinespace[1pt]
      PCGrad & $\tau > 1$ & 21.30 & 7.00 & 26.47 & 34.90 & 20.78 & 33.14 & 23.93 & +0.96 \\
      MFTCoder & $\tau > 1$ & 15.16 & 4.80 & 28.36 & 33.93 & 21.51 & 31.13 & 22.48 & -0.49 \\
      CoBa & $\tau > 1$ & 16.91 & 6.00 & 25.69 & 33.44 & 21.89 & 33.08 & 22.84 & -0.13 \\
      FAMO & $\tau > 1$ & 18.88 & 6.20 & 27.73 & \textbf{36.36} & \textbf{23.86} & 32.03 & 24.18 & +1.21 \\
      \midrule
      \addlinespace[2pt]
      \multicolumn{10}{c}{\textbf{Our Methods}} \\
      \addlinespace[1pt]
      Fixed-Bi. & $\tau > 1$ & \textbf{23.65} & 8.80 & \textbf{32.05} & 30.19 & 17.25 & 31.57 & 23.92 & +0.95 \\
      Fixed-Uni. & $\tau > 1$ & 20.55 & \textbf{9.20} & 29.77 & 33.60 & 23.82 & 30.11 & 24.51 & +1.54 \\
      Adaptive & $\tau > 1$ & 22.44 & 7.80 & 26.87 & 34.09 & 21.76 & \textbf{34.29} & \textbf{24.54} & \textbf{+1.57} \\
      \bottomrule
    \end{tabular}%
    }
\end{table}

We rerun the prior single-model baselines listed in the table under the same
temperature-smoothed sampling distribution ($\tau>1$) as the reported
\method{} branch. Smoothed SFT serves as the sampling-matched reference.

Under this matched setting, FAMO is the strongest prior baseline at 24.18.
The smoothed-sampling branch of Adaptive achieves the highest macro-average
in the table at 24.54, improving over Smoothed SFT by 1.57 points and FAMO
by 0.36 points. Fixed-Uni reaches 24.51, close to Adaptive. The advantage
over the evaluated prior baselines therefore persists when their sampling
distribution is matched to that of the reported AMD branch.

\section{Sensitivity to Fixed Distillation Weights}
\label{app:fixed_weight_sweep}
\label{sec:analysis_sensitivity_compute}

\begin{figure}[htbp]
  \centering
  \begin{minipage}[t]{0.45\linewidth}
    \centering
    \includegraphics[width=\linewidth]{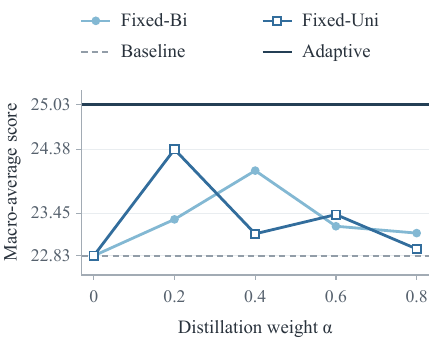}
    \small (a) Empirical sampling ($\tau=1$)
  \end{minipage}\hspace{0.02\linewidth}
  \begin{minipage}[t]{0.45\linewidth}
    \centering
    \includegraphics[width=\linewidth]{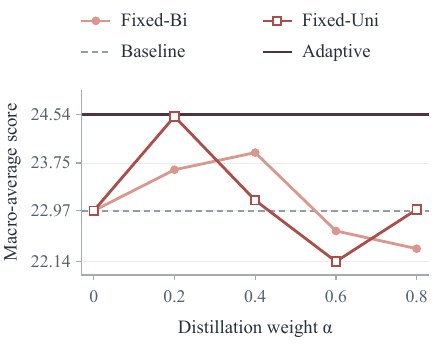}
    \small (b) Smoothed sampling ($\tau=3$)
  \end{minipage}
  \caption{Sensitivity to fixed distillation weights on Llama-3.2-1B.
  Scores are six-benchmark macro-averages. Dashed and solid horizontal lines
  indicate the SFT baseline and Adaptive, respectively, for each sampling branch.}
  \label{fig:weight_sweep_pair_figures}
\end{figure}

Figure~\ref{fig:weight_sweep_pair_figures} compares Adaptive with a sweep of
fixed weights for Fixed-Bi and Fixed-Uni in the empirical- and smoothed-sampling
branches on Llama-3.2-1B.
Fixed-Uni achieves its highest plotted score at $\alpha=0.2$ in both branches.
For $\tau=1$, Adaptive scores 25.03 versus the best fixed-weight result of
24.38, a gain of 0.65 points. For $\tau=3$, Adaptive and Fixed-Uni score
24.54 and 24.51, respectively, a difference of 0.03 points.
The gain from adaptation is therefore more pronounced in the empirical-sampling
branch.

\section{Weight-Update Schedule Sensitivity}
\label{app:scheduler_sensitivity}

\begin{table}[H]
    \caption{Sensitivity of the Llama-3.2-1B empirical-sampling model to the
  step-size schedule $\mu$. Scores are six-benchmark macro-averages; bold
  marks the best score.}
    \label{tab:scheduler_ablation}
  \centering

  \small
  \setlength{\tabcolsep}{6pt}
  \renewcommand{\arraystretch}{1.35}
  \begin{tabular}{c c c}
    \toprule
    \textbf{\#} & \textbf{Schedule} & \textbf{Macro Avg.} \\
    \midrule
    1 & $\displaystyle \mu = 1$ & 23.96 \\
    \cmidrule{1-3}
    2 & $\displaystyle \mu = \sqrt{\frac{T_{\max} - 0.8\,t}{T_{\max}}}$ & 24.89 \\
    \cmidrule{1-3}
    3 & $\displaystyle \mu = \sqrt{\max\left(0,\frac{T_{\max} - 1.2\,t}{T_{\max}}\right)}$ & 23.64 \\
    \cmidrule{1-3}
    4 & $\displaystyle \mu = \sqrt{\frac{T_{\max} - t}{T_{\max}}}$ & \textbf{25.03} \\
    \bottomrule
  \end{tabular}
\end{table}

Table~\ref{tab:scheduler_ablation} compares step-size schedules using
the empirical-sampling branch. The standard square-root decay achieves 25.03,
followed by slower decay at 24.89, a constant step at 23.96, and faster decay
at 23.64. Both the standard and slower schedules outperform the constant and
faster-decaying alternatives, supporting a gradual reduction in update
magnitude.

\section{Longer-Budget Stress Test}
\label{app:longer_budget_stress}

\begin{table}[H]
    \caption{Longer-budget stress test on Llama-3.2-1B. Single-model baselines
  train for six epochs, twice the three-epoch main protocol used by AMD.
  For each AMD regime, we report the higher-scoring model from the
  collaborative run. Scores are six-benchmark macro-averages.}
  \label{tab:compute_budget_stress_test}
  \centering

  \small
  \setlength{\tabcolsep}{9pt}
  \renewcommand{\arraystretch}{1.08}
  \begin{tabular}{@{}lc@{}}
    \toprule
    \textbf{Method} & \textbf{Macro Avg.} \\
    \midrule
    \multicolumn{2}{@{}l}{\textit{Single-model baselines, six epochs}} \\
    \quad Multi-task SFT  & 23.05 \\
    \quad Smoothed SFT  & 22.94 \\
    \quad PCGrad & 23.50 \\
    \quad FAMO & 24.16 \\
    \quad CoBa & 23.35 \\
    \quad MFTCoder & 23.52 \\
    \quad HBO & 24.11 \\
    \addlinespace[2pt]
    \midrule
    \multicolumn{2}{@{}l}{\textit{AMD regimes, main protocol}} \\
    \quad Fixed-Bi & 24.07 \\
    \quad Fixed-Uni & 24.51 \\
    \quad Adaptive & \textbf{25.03} \\
    \bottomrule
  \end{tabular}
\end{table}

Table~\ref{tab:compute_budget_stress_test} evaluates single-model baselines
trained for six epochs alongside the higher-scoring branch of each \method{}
regime trained under the three-epoch main protocol. FAMO achieves the highest
single-model score of 24.16, compared with 24.07 for \biregime{}, 24.51 for
\uniregime{}, and 25.03 for Adaptive. Adaptive retains the highest score when
the single-model baselines receive twice as many training epochs. These epoch
budgets correspond to different computational costs because \method{} also
uses cross-model forward passes and, in Adaptive, training probes and
validation.

Adaptive performs five weight updates during training, each with three probes
using 1\% of a one-epoch training budget. The probes therefore add 0.15
epoch-equivalents of two-model training, or 5\% of the regular three-epoch
training budget. This estimate excludes validation and checkpoint-restoration
overhead.

\end{document}